\pdfoutput=1

\documentclass[11pt]{article}

\usepackage{acl}

\usepackage{times}
\usepackage{latexsym}
\usepackage{multirow}
\usepackage{graphicx}
\usepackage{amsmath}
\usepackage{amsfonts}

\usepackage[T1]{fontenc}

\usepackage[utf8]{inputenc}

\usepackage{microtype}

\title{Modeling Multi-Granularity Hierarchical Features for Relation Extraction}

\author{Xinnian Liang\textsuperscript{1}\footnotemark[1], Shuangzhi Wu\textsuperscript{2}, Mu Li\textsuperscript{2} \and Zhoujun Li\textsuperscript{1}\footnotemark[2]\\ 
  \textsuperscript{1}State Key Lab of Software Development Environment, Beihang University, Beijing, China \\
  \textsuperscript{2}Tencent Cloud Xiaowei, Beijing, China\\
  \texttt{\{xnliang,lizj\}@buaa.edu.cn};
  \texttt{frostwu@tencent.com,limugx@qq.com};\\
}

\begin{document}
\maketitle

\renewcommand{\thefootnote}{\fnsymbol{footnote}}
\footnotetext[1]{Contribution during internship at Tencent Inc.} 
\footnotetext[2]{Corresponding Author}
\renewcommand{\thefootnote}{\arabic{footnote}}

\begin{abstract}
Relation extraction is a key task in Natural Language Processing (NLP), which aims to extract relations between entity pairs from given texts. Recently, relation extraction (RE) has achieved remarkable progress with the development of deep neural networks. Most existing research focuses on constructing explicit structured features using external knowledge such as knowledge graph and dependency tree. In this paper, we propose a novel method to extract multi-granularity features based solely on the original input sentences. We show that effective structured features can be attained even without external knowledge. Three kinds of features based on the input sentences are fully exploited, which are in entity mention level, segment level, and sentence level. All the three are jointly and hierarchically modeled. We evaluate our method on three public benchmarks: SemEval 2010 Task 8, Tacred, and Tacred Revisited. To verify the effectiveness, we apply our method to different encoders such as LSTM and BERT. Experimental results show that our method significantly outperforms existing state-of-the-art models that even use external knowledge. Extensive analyses demonstrate that the performance of our model is contributed by the capture of multi-granularity features and the model of their hierarchical structure. Code and data are available at \url{https://github.com/xnliang98/sms}.
\end{abstract}

\section{Introduction}
Relation extraction (RE) is a fundamental task in Natural Language Processing (NLP), which aims to extract relations between entity pairs from given plain texts. RE is the cornerstone of many downstream NLP tasks, such as knowledge base construction \cite{ji-grishman-2011-knowledge}, question answering \cite{yu-etal-2017-improved}, and information extraction \cite{fader-etal-2011-identifying}.

Most recent works focus on constructing explicit structured features using external knowledge such as knowledge graph, entity features and dependency tree.
To infuse prior knowledge from existing knowledge graph, recent works \cite{peters-etal-2019-knowledge,wang2020kepler,wang2020kadapter} proposed some pre-train tasks to help model learn and select proper prior knowledge in the pre-training stage.
\citet{baldini-soares-etal-2019-matching,yamada-etal-2020-luke,peng-etal-2020-learning} force model learning entitiy-related information via well-designed pre-train tasks.
\citet{zhang2018graph,guo2019attention,xue2020gdpnet,chen2020efficient} encode dependency tree with graph neural network \cite{kipf2016semi} (GNN) to help RE models capture non-local syntactic relation. All of them achieve a remarkable performance via employing external information from different structured features.

However, they either need time-consuming pre-training with external knowledge or need an external tool to get a dependency tree which may introduce unnecessary noise. 
In this paper, we aim to attain effective structured features based solely on the original input sentences.
To this end, we analyze previous typical works and find that three kinds of features mainly affect the performance of RE models, which are entity mention level\footnote{entity mentions contain the entity itself and co-references of it.}, segment\footnote{continuous words in sentence (n-gram)} level and sentence level features.
Sentence level and entity mention \cite{baldini-soares-etal-2019-matching,yamada-etal-2020-luke,peng-etal-2020-learning} level features were widely used by previous works but segment level feature \cite{yu2019beyond,joshi-etal-2020-spanBERT} does not get as much attention as the previous two features. 
These three level features can provide different granularity information from input sentences for relation prediction \cite{chowdhury-lavelli-2012-combining,Kim2010}. 
However, recent works did not consider them at the same time and ignored the structure and interactive of them.

\begin{figure}[]
	\centering
	\includegraphics[scale=0.6]{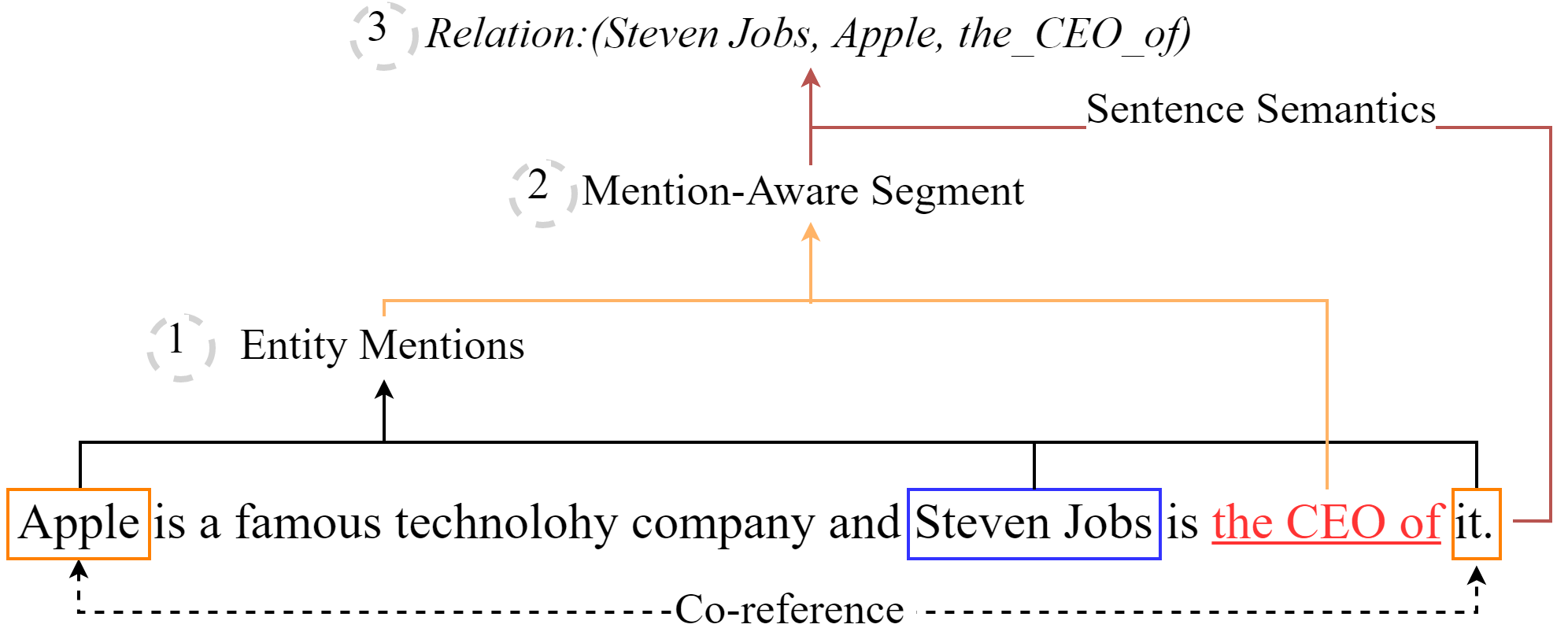}
	\caption{An Example for relation extraction which shows the hierarchical structure between entity mention level and segment level features.}
	\label{fig:example}
\end{figure}
We employ a simple example in Figure \ref{fig:example} to show the hierarchical and joint structure of the previous three granularities features.
The hierarchical structure is between mention level and segment level features.
This example gives sentence and entity pairs (``\textit{Steven Jobs}", ``\textit{Apple}"). 
We can find that the relation ``\textit{the\_CEO\_of}" of given entity ``\textit{Apple}" and entity ``\textit{Steven Jobs}" is built upon the core segment ``\textit{the CEO of}" between the entity mention ``\textit{it}" (i.e. co-reference of entity ``\textit{Apple}") and entity ``\textit{Steven Jobs}".
To extract relation from this example, RE models need to first capture mention level features of given entities and then catch core segment level feature ``\textit{the CEO of}" which is related to mention level features.
Finally, RE models can easily predict the relation with previous two-level hierarchical features.
Besides, sentence-level semantic features can assist RE models to predict the relation of examples without an explicit pattern of entity mentions and segments.

Following previous intuitive process, we propose a novel method which extracts multi-granularity features based solely on the original input sentences.
Specifically, we design a hierarchical mention-aware segment attention, which employs a hierarchical attention mechanism to build association between entity mention level and segment level features. Besides, we employ a global semantic attention to get a deeper understanding of sentence level features from input sentence representation. Finally, we aggregate previously extracted multi-granularity features with a simple fully-connected layer to predict the relation.

To evaluate the effectiveness of our method, we combine our method with different text encoders (e.g. LSTM and BERT) and results show that our method can bring significant improvement for all of them.
Compared with models without using external knowledge, SpanBERT with our method can achieve a new state-of-the-art result on Semeval 2010 Task 8, Tacred and Tacred Revisited. It is deserved to mention that the performance of our model is very competitive with the state-of-the-art models, which employ large-scale extra training data or information. We also do many analyses to demonstrate that features from representation of input itself are enough for the sentence-level RE tasks and multi-granularity features with hierarchical structure are crucial for relation prediction.



\section{Methodology}
\begin{figure*}[]
	\centering
	\includegraphics[scale=0.6]{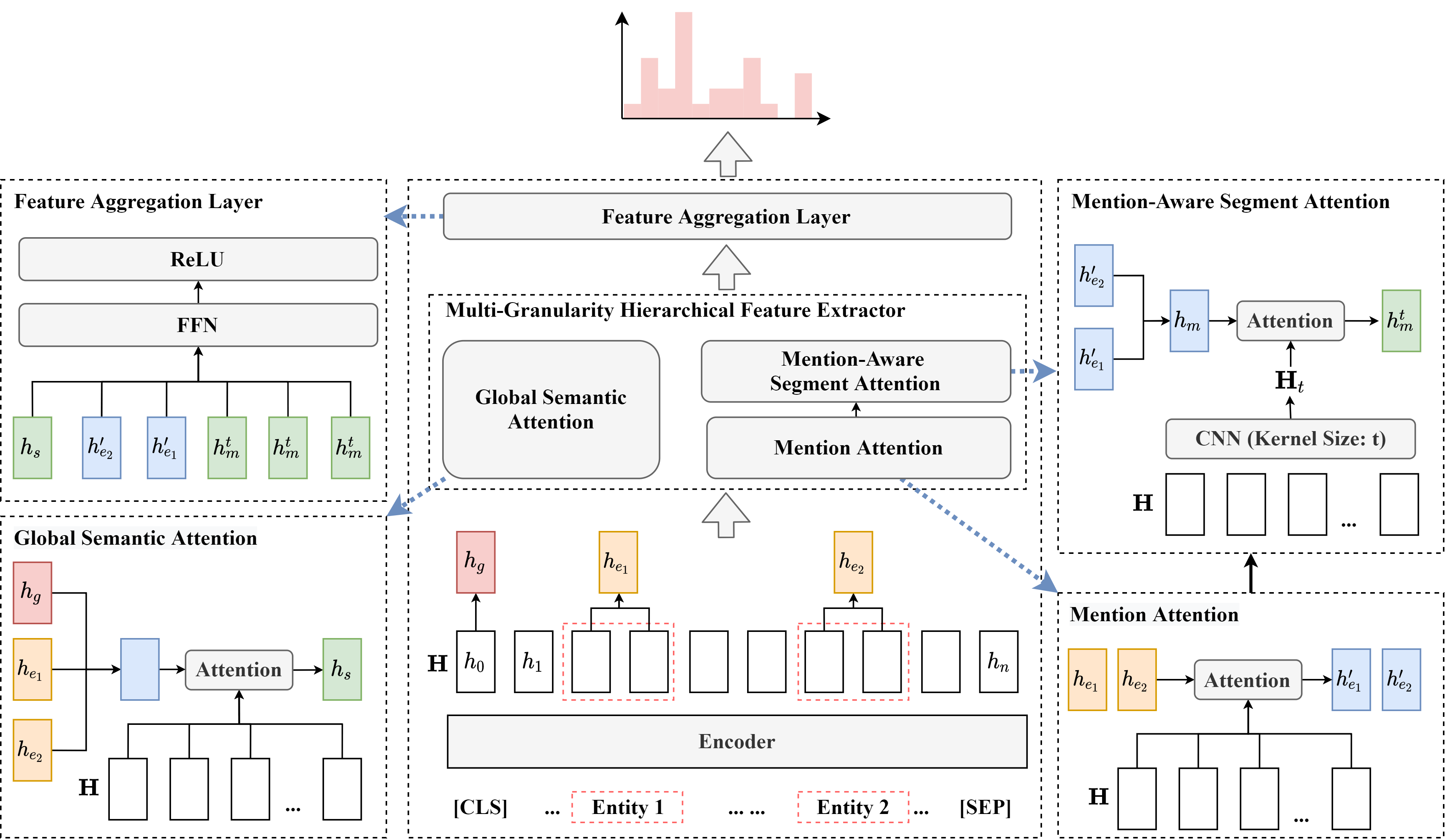}
	\caption{\textbf{Middle: }The structure of our proposed multi-granularity hierarchical feature extractor. \textbf{Left:} Details of global semantic attention (sentence level feature) and feature aggregation layer. \textbf{Right:} Details of mention attention (entity mention level feature) and mention-aware segment attention (segment level feature).}
	\label{fig:model}
\end{figure*}
The structure of our model and details of each component is shown in figure \ref{fig:model}. We can see the overall architecture in the middle. It is divided into three components from bottom to top: 1) A text encoder which is employed to obtain text vector representations; 2) A multi-granularity hierarchical feature extractor which can exploit effective structured features from text representations; 3) A feature aggregation layer which aggregate previous multi-granularity features for relation prediction. 
In this section, we will introduce details of three components.

Firstly, we formalize the relation extraction task. Let $\mathtt{x} = \{x_1, x_2, ..., x_n \}$ be a sequence of input tokens, where $x_0 = [\mathtt{CLS}]$ and $x_n = [\mathtt{SEP}]$ are special start and end tokens for BERT-related encoders.
Let $s_1 = (i, j)$ and $s_2 = (k, l)$ be pairs of entity indices. The indices in $s_1$ and $s_2$ delimit entities in $\mathtt{x}$: $[x_i, \dots, x_{j-1}]$ and $[x_k, \dots, x_{l-1}]$.
Our goal is to learn a function $P(r) = f_\theta(\mathtt x, s_1, s_2)$, where $r \in \mathcal{R}$ indicates the relation between the entity pairs, which is marked by $s1$ and $s2$. $\mathcal{R}$ is a pre-defined relation set.

\subsection{Encoder Layer}
We first employ a text encoder (e.g. BERT) to map tokens in input sentences into vector representations which can be formalized by Equ. (\ref{eq1}).
\begin{equation}
    \mathbf{H} = \{h_0, \dots, h_n\} = f_{encoder}(x_0, \dots, x_n)
    \label{eq1}
\end{equation}
Where $\mathbf{H}= \{h_0, \dots, h_n\}$ is the vector representation of input sentences. 

Our work is built upon $\mathbf{H}$ and does not need any external information.
We employ a max-pooling operation to obtain shallow features of entity pairs and input sentences. $h_{e_1} = \mathtt{Maxpooling}(h_{i:j})$ and $h_{e_2} = \mathtt{Maxpooling}(h_{k:l})$ are the representations of entity pairs. $h_g = \mathtt{Maxpooling}(\mathbf{H})$ is the vector representation of input sentences which contains global semantic information.

\subsection{Multi-Granularity Hierarchical Feature Extractor}
The multi-granularity hierarchical feature extractor is the core component of our method and it consists of three attention mechanism for different granularity features extraction: 1) mention attention which is designed to entity mention features of given entity pairs; 2) mention-aware segment attention which is based on the entity mention features from previous mention attention and aim to extract core segment level feature which is related to entity mentions; 3) global semantic attention which focuses on the sentence level feature.

\subsubsection{Mention Attention}
The structure of mention attention is shown in the right bottom of Figure \ref{fig:model}.
To capture more information about given entity pairs from input sentences, we extract entity mention level features by modeling the co-references (mentions) of entities implicitly. 
We employ a mention attention to capture information about entity 1 and 2 respectively. Specifically, we can use the representation of an entity as a query to obtain the entity mention feature from $\mathbf H$ by Equ. (\ref{eq3}).
\begin{equation}
    \begin{aligned}
        h'_{e_1} &= \mathtt{Softmax}(\frac{\mathbf{H} \cdot h_{e_1}}{\sqrt{d}}) \cdot \mathbf{H}  \\
        h'_{e_2} &= \mathtt{Softmax}(\frac{\mathbf{H} \cdot h_{e_2}}{\sqrt{d}}) \cdot \mathbf{H}
    \end{aligned}
\label{eq3}
\end{equation}
Where $d$ is the dimension of vector representation and used to normalize vectors. 
Then, $h'_{e_1}$ and $h'_{e_2}$ model the mentions of given entity pairs implicitly and contain more entity semantic information than $h_{e_1}$ and $h_{e_2}$.

\subsubsection{Mention-Aware Segment Attention}
The structure of mention-aware segment attention is shown in the right top of Figure \ref{fig:model}. And the mention-aware segment attention is a hierarchical structure based on the entity mention features $h'_{e_1}$ and $h'_{e_2}$ from mention attention.

Before introducing mention-aware segments attention, we first introduce how to get the representations of segments.
We employ convolutional neural networks (CNN) with different kernel sizes to obtain all n-gram segments in texts, which can effectively capture local n-gram information with Equ. (\ref{eq4}).
\begin{equation}
    \mathbf{H}_t = \mathtt{CNN}_t(\mathbf{H}), t \in \{1, 2, 3\}
    \label{eq4}
\end{equation}
Where $t$ is the kernel size of CNN and is empirically set as $t \in \{1, 2, 3\}$ which means extract 1-gram, 2-gram ,and 3-gram segment level features.

Intuitively, the valuable segments should be highly related to given entity pairs, which can help the model to decide the relation of given entity pairs. 
Entity mention features $h'_{e_1}$ and $h'_{e_2}$ contain comprehensive information of given entity pairs and $\mathbf{H}_t$ contain 1,2,3-gram segment level features. 
We can extract mention-aware segment level features by simply linking them with attention mechanisms by Equ. (\ref{eq5}).
\begin{equation}
    h_m^t = \mathtt{Softmax}(\frac{\mathbf{H}_t \cdot (W_m[h'_{e_1}; h'_{e_2}])}{\sqrt{d}}) \cdot \mathbf{H}_t
    \label{eq5}
\end{equation}
Then, we get $\{h_m^t\}_{t=1,2,3}$ which capture different granularity segments features. 

\subsubsection{Global Semantic Attention}
The structure of global semantic attention is shown in the left bottom of Figure \ref{fig:model}.
Previous works always directly concatenate vector representation $[h_{e_1}; h_{e_2}; h_g]$ as the global semantic feature of input text. We argue this is not enough to help model capture deeper sentence level semantic information for RE.
Different from them, to obtain better global sentence-level semantic feature, we employ an attention operation called global semantic attention which use the concatenation of $[h_{e_1}; h_{e_2}; h_g]$ as query to capture deeper semantic feature from context representation $\mathbf H$ by Equ. (\ref{eq2}).
\begin{equation}
    h_s = \mathtt{Softmax}(\frac{\mathbf{H} \cdot (W_s[h_{e_1}; h_{e_2}; h_g])}{\sqrt{d}}) \cdot \mathbf{H}
    \label{eq2}
\end{equation}
Where $W_s \in \mathbb{R}^{d \times 3d}$ is a linear transform matrix, and $d$ is a hidden dimension of vectors.
The concatenation of $[h_{e_1}; h_{e_2}; h_g]$ is used as a query of the attention operation, which can force the extracted global semantic representation $h_s$ contain entity mention related sentence level feature.

\subsection{Feature Aggregation Layer}
The structure of the feature aggregation layer is shown in the left top of Figure \ref{fig:model}.
We aggregate previous multi-granularity features by Equ. (\ref{eq6}).
\begin{equation}
    h_o = \mathtt{ReLU}(W_a[h_{s};h'_{e_1};h'_{e_1};h_m^1;;h_m^2;;h_m^3])
    \label{eq6}
\end{equation}
Where $W_a \in \mathbb R^{6d \times d}$ is a linear transform matrix and $\mathtt{ReLU}$ is a nonlinear activation function.
\subsection{Classification}
Finally, we employ a softmax function to output the probability of each relation label as follows:
\begin{equation}
P(r | \mathtt{x}, s_1, s_2) =\mathtt{Softmax}(W_{o}h_o)
\end{equation}
The whole model is trained with cross entropy loss function.
We call the multi-granularity hierarchical feature extractor: SMS (relation extraction with \textbf{S}entence level, \textbf{M}ention level and mention-aware \textbf{S}egment level features).

\section{Experiments}
\begin{table}[]
    \centering
    \small
    \begin{tabular}{lcc}
    \hline
     & Tacred & Semeval \\ \hline
    lr & 3e-5 & 2e-5 \\
    warmup steps & 300 & 0 \\
    batch size & 64 & 32 \\
    V100 GPU & 4x & 1x \\
    epochs & 4 & 10 \\
    max length & 128 & 128 \\ \hline
    \end{tabular}
    \caption{Hyper-parameters used in training.}
    \label{tab:params}
\end{table}
\begin{table*}[]
\centering
\small
\begin{tabular}{clcccccc}
\hline
& {\multirow{2}{*}{Model}} & \multicolumn{3}{c}{Tacred}           & \multicolumn{3}{c}{Tacred Revisited} \\
\multicolumn{2}{c}{}                 & P($\Delta P$)    & R($\Delta R$)    & F1($\Delta F1$)   & P($\Delta P$)    & R($\Delta R$)    & F1($\Delta F1$)    \\ \hline
\multirow{5}{*}{1}   & LSTM                & 62.5        & 63.4      & 62.9       & 67.7       & \textbf{73.1}       & 70.3       \\
                    & PA-LSTM*        & 65.7 & 64.5 & 65.1 & 73.6 & 72.8 & 73.2 \\
                    & SA-LSTM        & 68.1 & \textbf{65.7} & 66.9 &\textbf{ 78.3} & 72.5 & \textbf{75.4 }\\
                    & C-GCN*          & 69.9 & 63.3 & 66.4 & 76.8 & 71.4 & 74.1 \\
                    & C-AGGCN*        & \textbf{71.9} & 63.4 & \textbf{67.5} & 78.2 & 70.5 & 74.3 \\
                     \hline
\multirow{7}{*}{2} & TRE            & -    & -    & 67.4 & -    & -    & 75.3 \\
                    & BERT-base      & 68.1 & 67.7 & 67.9 & 69.4 & 75.8 & 72.6 \\
                    & BERT-large     & 69.2 & 69.4 & 69.3 & 75.1 & 74.8 & 75.0 \\
                    & BERT+LSTM      & 73.3 & 63.1 & 67.8 & 74.1 & 73.9 & 74.0 \\
                    & SpanBERT-base  & 67.6 & 68.6 & 68.1 & 69.1 & 78.2 & 73.7 \\
                    & SpanBERT-large & 70.8 & 70.9 & 70.8 & 77.8 & 78.3 & 78.0 \\
                    & DG-SpanBERT-large*   &\textbf{ 71.4}        & \textbf{71.6}      & \textbf{71.5}       & \textbf{79.2}       &\textbf{ 78.6}       & \textbf{78.9  }     \\
                     \hline
\multirow{5}{*}{3}   
                    & MTB\dag            & -    & -    & 71.5 & -    & -    & -    \\
                    & KnowBERT-W+W\ddag   & \textbf{71.6} & 71.4 & 71.5 & 79.0 & 79.6 & 79.3 \\
                    & KEPLER\ddag      & 70.4 & 73.0 & 71.7 &    -  &    -  &   -       \\
                    & K-Adapter\ddag      & 70.1 & 74.0 & 72.0 &   -   &    -  &  -       \\
                    & LUKE\dag           & 70.4 & \textbf{75.1} & \textbf{72.7} &  \textbf{79.7}   &  \textbf{80.6}    &   \textbf{80.2}      \\ \hline
\multirow{5}{*}{4}
                    & LSTM+SMS            & \textbf{\underline{72.8(+10.3)}} & 64.6(1.2) & 68.4(+5.5) & 80.8(+3.1) & 72.2(-0.9) & 75.9(+5.6) \\
                    &SpanBERT-base+SMS  & 72.6(+5.0) & 68.4(-0.2) & 70.5(+2.4) & 79.1(+10.0) & 77.7(-0.5)  & 78.3(+4.6)        \\
                    &SpanBERT-large+SMS & 72.2(+1.4) & \textbf{\underline{71.6(+0.7)}} & \textbf{\underline{71.9(+1.1)}} & \textbf{\underline{79.3(+1.5)}} & \textbf{\underline{80.4(+2.1)}} & \textbf{\underline{79.8(+1.8)}}       \\
                    &BERT-base+SMS      & 69.4(+1.3) & 70.2(+2.5) & 69.7(+1.8) & 77.0(+7.6) & 80.1(+4.3) & 78.5(+5.9)        \\
                    &BERT-large+SMS     & 70.7(+1.5) & 69.1(-0.3) & 69.9(+0.6) & 78.9(+3.8) & 79.2(+4.4) & 79.1(+4.1)        \\ \hline
\end{tabular}
\caption{Results on Tacred and Tacred Revisited. Bold means the best results in each block. Underline means the best results in block 1, 2, and 4. * means that the model employs dependency tree information. \dag means that the model employs extra training data to pre-train the model. \ddag means the model employs knowledge graphs.}
\label{tab:res_main}
\end{table*}
\subsection{Datasets}

We evaluate the performance of our method on Semeval 2010 Task 8, Tacred and Tacred Revisited datasets. 

\textbf{SemEval 2010 Task 8} \cite{hendrickx-etal-2010-semeval} is a public dataset which contains 10,717 instances with 9 relations. The training/validation/test set contains 7,000/1,000/2,717 instances respectively.

\textbf{Tacred}\footnote{https://nlp.stanford.edu/projects/tacred/} is one of the largest, most widely used crowd-sourced datasets for Relation Extraction (RE), which is introduced by \cite{zhang2017position}, with 106,264 examples built over newswire and web text from the corpus used in the yearly TAC Knowledge Base Population (TAC KBP) challenges. The training/validation/test set contains 68,124/22,631/15,509 instances respectively.
It covers 42 relation types including 41 relation types and a \textit{no\_relation} type and contains longer sentences with an average sentence length of 36.4. 

\textbf{Tacred Revisited}\footnote{https://github.com/DFKI-NLP/tacrev} was proposed by \cite{DBLP:conf/acl/AltGH20a} which aims to improve the accuracy and reliability of future RE method evaluations. 
They validate the most challenging 5K examples in the development and test sets using trained annotators and find that label errors account for 8\% absolute F1 test error, and that more than 50\% of the examples need to be relabeled.
Then, they relabeled the test set and released the Tacred Revisited dataset.

\subsection{Settings}

The setting of hyper-parameters is shown in table \ref{tab:params}. Following the implementation details mentioned in \cite{zhang2017position}, we employ the “entity mask” strategy and the “multi-channel” strategy during experiments. 
The former means replacing each subject entity (and object entity similarly) in the original sentence with a special \texttt{[SUBJ-$\langle$NER$\rangle$]} token. 
All of our reported results are the mean of 5 results with different seeds, which are randomly selected. We evaluate the models on Tacred with the official script\footnote{https://github.com/yuhaozhang/tacred-relation} in terms of the Macro-F1 score and on Semeval with the official script \textit{semeval2010\_task8\_scorer-v1.2.pl}.

When employing LSTM as the encoder, we employ a single-layer bidirectional LSTM with the hidden dimension size set to 200, we set dropout after the input layer and before the output layer with $p$ = 0.5. We use stochastic gradient descent (SGD) with epochs of 30, learning rate of 1.0, decay weight of 0.5 and batch sizes of 50 to train the model. 
The latter is to augment the input by concatenating it with part-of-speech (POS) and named entity recognition (NER) embeddings.
Glove \cite{pennington2014glove} embedding with 300-dimension is used for initializing word embedding layers in LSTM+SMS. 
NER embedding, POS embedding and position embedding are randomly initialized with 30-dimension vectors from uniform distribution. 

\subsection{Comparison Models}
We mainly compare with models which are based on pre-trained language models (e.g. BERT).
We reproduce the results of \textbf{BERT} and \textbf{SpanBERT} to evaluate the improvement of our method. We also compared other models with pre-trained language models.
\textbf{TRE} \cite{alt_improving_2019}, which uses the unidirectional OpenAI Generative Pre-Trained Transformer (GPT) \cite{radford2019language}.
\textbf{BERT-LSTM} \cite{shi2019simple}, which stacks bidirectional LSTM layer to BERT encoder.
\textbf{DG-SpanBERT}, which replaced the encoder of C-GCN \cite{zhang2018graph} with SpanBERT and achieved the new state-of-the-art result without extra training data.
\textbf{MTB} \cite{baldini-soares-etal-2019-matching}, which incorporates an intermediate ``matching the blanks'' pre-training on the entity-linked text based on English Wikipedia.
\textbf{KnowBERT-W+W} \cite{Peters2019KnowledgeEC}, which is an an advanced version of KnowBERT.
\textbf{KEPLER} \cite{wang2020kepler}, which integrates factual knowledge with the supervision of the knowledge embedding objective.
\textbf{K-Adapter} \cite{wang2020kadapter}, which consists of a RoBERTa model and an adapter to select adaptive knowledge.
\textbf{LUKE} \cite{yamada-etal-2020-luke}, which is trained with a new pre-training task which involves predicting randomly masked words and entities in a large entity-annotated corpus retrieved from Wikipedia and introduce a new entity-aware attention mechanism.

In order to further prove the effectiveness of our SMS, we use bi-directional LSTM as encoder, and compare with models which do not use pre-trained language models. 
We choose two sequence-based models.
\textbf{PA-LSTM} \cite{zhang2017position}, which employ Bi-LSTM to encoder the plain text and combine with position-aware attention mechanism to extract relation. PA-SLTM is the benchmark of Tacred. 
\textbf{SA-LSTM} \cite{yu2019beyond}, which employ CRF to learn segment-level attention and is the best sequence-based model of Tacred.

We also compare our model with two other dependency-based models which make use of GCN \cite{kipf2016semi} to capture semantic information from the dependency tree. 
\textbf{C-GCN} \cite{zhang2018graph}, which applies pruning strategy and GCN to extract features from tree structure for relation extraction.
\textbf{C-AGGCN} \cite{guo2019attention}, which introduces self-attention to build a soft adjacent matrix as input of Dense GCN to learn tree structure features.

\subsection{Results on Tacred and Tacred Revisited}

We first report the results or our model on Tacred and Tacred Revisited on Table \ref{tab:res_main}.
Compared models are divided into three categories: 1) models with Bi-LSTM encoder in block 1; 2) models with pre-trained models in block 2; 3) models with external knowledge in block 3. The results of our model are reported in block 4. 
We use * to mark models with dependency trees which are obtained with external tools. 
We use \dag to mark models which use external training data to pre-train the model and \ddag to mark models which employ knowledge graphs to pre-train or fine-tune the model. Models with \dag and \ddag require external data and we do not directly compare them.

\subsubsection{Compare with Pre-trained models}
We can see that our SMS can bring at least 0.6 and up to 5.5 F1 score improvement for the original encoder on Tacred dataset.
On the Tacred Revisited dataset, our SMS can bring at least 1.8 and up to 5.9 F1 score improvement for the original encoder.
Overall, different encoders with SMS all can obtain remarkable improvement on both datasets. This proves that our SMS really captures effective features from input sentence representations, which can not get directly from the representations.
Compared with models which employ pre-trained models without external knowledge (i.e. training data or knowledge graph) in block 2, pretrained models with our SMS in block 4 overall perform better and SpanBERT-large+SMS achieve new state-of-the art results on both datasets.
In addition, we can see that the performance of SpanBERT-large+SMS is better than MTB, KnowBERT-W+W, and KEPLER in block 3 and is competitive with K-Adapter and LUKE.

The increase of F1 score is more conspicuous on Tacred Revisited compared with Tacred. This phenomenon is further evidence that existing models have neared the upper limit of Tacred, which have many mislabeled examples. Besides, we can see that models based on SpanBERT all have a pretty good performance. This phenomenon proves the importance of segment level features.

\subsubsection{Compare with LSTM-based models}
To further evaluate the effectiveness of our method SMS, we specially combine SMS with LSTM encoder. We can observe that our model also outperforms the model with LSTM encoder in block 1. Dependency-based models with graph neural networks (C-GCN and C-AGGCN) have a remarkable performance on Tacred and models which focus on segments (SA-LSTM) have a better performance on Tacred Revisited. This phenomenon means that directly modeling the segment level feature can not effectively overcome the noise from mislabeled examples and the introduction of graph structure with dependency trees can help models tackle some influence from wrong examples in the dataset itself.

However, our LSTM+SMS can outperform them on both datasets due to our mention-aware segment attention can alleviate influence from mislabeled entity pairs via modeling entity mention level feature and hierarchical structure.

\subsection{Results on SemEval 2010 Task 8}
\begin{table}[]
\centering
\small
\begin{tabular}{lc}
\hline
\multirow{2}{*}{Models}& SemEval \\
& F1($\Delta F1$) \\ \hline
LSTM               &  82.7 \\
LSTM+Attention     &  84.0 \\ \hline
TRE                &  87.1 \\
BERT-base          &  87.9 \\
BERT-large         &  88.8 \\
SpanBERT-base      &  88.2\\
SpanBERT-large     &  89.4 \\
R-BERT \cite{wu-etal-2019-r-bert}     &  89.3 \\ \hline
MTB \dag               &  89.5 \\
KnowBert-W+W \ddag       &  89.1 \\ \hline
LSTM+SMS           &  86.8(+4.1) \\
BERT-base+SMS      &  88.4(+0.5) \\
BERT-large+SMS     &  89.8(+1.0) \\
SpanBERT-base+SMS  &  88.5(+0.3) \\
SpanBERT-large+SMS &  \textbf{90.3(+0.9)} \\ \hline
\end{tabular}
\caption{Results on SemEval 2010 task8. \dag means that the model employs external knowledge to pre-train the model. \ddag means the model employs a knowledge base.}
\label{tab:res_semeval}
\end{table}
We also evaluate our SMS with different encoders on SemEval 2010 Task 8 dataset and results are reported in Tab. \ref{tab:res_semeval}.
We can observe that our SMS still brings remarkable improvement for different encoders, especially for LSTM encoders. SpanBERT-large+SMS outperforms all compared to strong baselines.
Besides, SpanBERT-large+SMS can beat models with external knowledge due to this dataset being simpler than Tacred which only has 9 relations and shorter input sentences. These reasons reduce the gain from the introduction of external knowledge.

We also can see that the improvement of LSTM with SMS is up to 4.1\% F1 score. We guess that pre-trained models contain a lot of semantic information from pre-training data which is similar to features from our SMS. However, LSTM only captures features from the plain texts and can achieve more improvement from our proposed SMS.

\section{Discussion}
\subsection{Ablation Study}
\begin{table}[]
	\centering
	\small
	\begin{tabular}{lc}
		\hline
		& F1($\Delta F1$) \\ \hline
		SpanBERT-large    & 78.0 \\
		\quad+ sentence level  & 78.6(+0.6) \\
		\quad+ mention level   & 78.8(+0.8) \\
		\quad+ segment level   & 79.4(+1.4) \\
	    \quad+ all    & \textbf{79.8(+1.8)} \\ \hline
	\end{tabular}
	\caption{Ablation study on Tacred Revisited test set.}
	\label{tab:ab}
\end{table}
To evaluate the contribution of each component of our SMS, we do an ablation study and results are shown in Tab. \ref{tab:ab}. 
We can observe that segment level features contribute the most for the F1 score, which are extracted by the mention-aware segment attention. This means the hierarchical structure between entity mention level and segment level feature really play a vital role for relation prediction.
In the future works, segment features need more attention.
We also can see that all three granularity features influence the performance obviously. 
This proves the capture of these three granularity features are proper for relation extraction tasks.

\subsection{Analysis with N-gram Segments}
\begin{figure}[]
    \centering
    \includegraphics[scale=0.45]{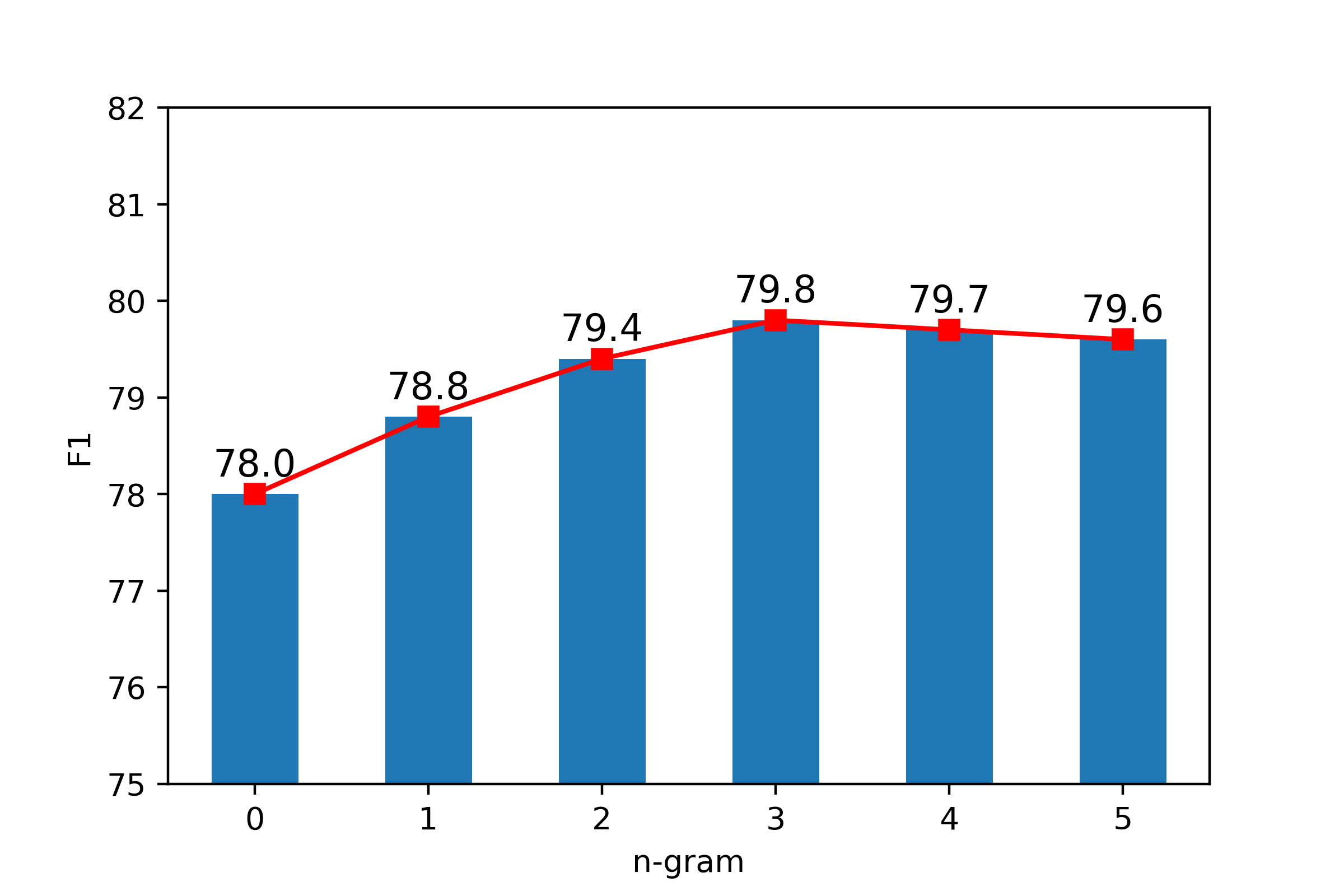}
    \caption{Performance on Tacred Revisited test set with different n-gram segments features. Number $n$ in x-axis means the model use $n$-gram segment features.}
    \label{fig:ngram}
\end{figure}
We show the performance on the Tacred Revisited test set with different n-gram segments features in Figure \ref{fig:ngram}. Number $n$ in the x-axis means the model uses $1-n$-gram segment features.   
We can observe that the model with 1,2,3-gram segment features achieves the best performance.  
Longer segment features can not bring improvement and may bring noise to the performance of the model.  
So we employ 1,2,3-gram segment level features in our paper.

\subsection{Case Study}
\begin{figure*}[]
    \centering
    \includegraphics[scale=0.9]{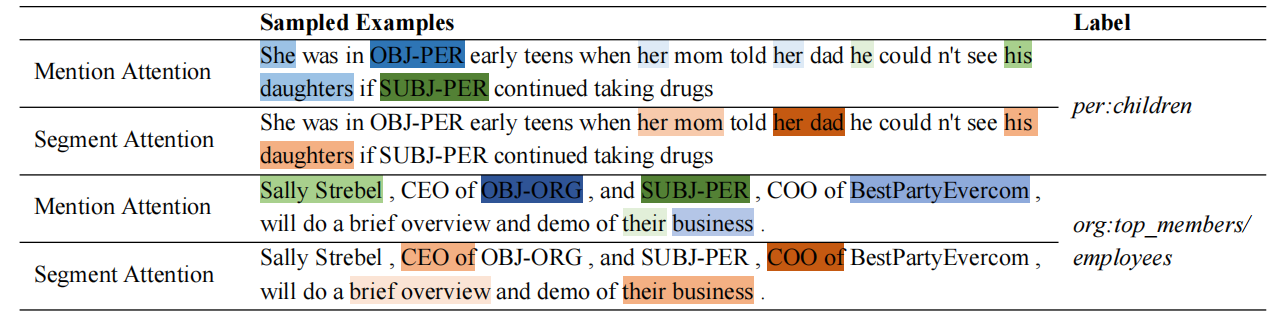}
    \caption{Two examples which are sampled from Tacred Revisited test set. The shade of the color represents how much attention is allocated. For the sake of perception, we did not color words with very low attention values.}
    \label{fig:case}
\end{figure*}
As shown in Figure \ref{fig:case}, we visualized the attention of our SMS with two examples which are sampled from Tacred test set. 
In the first example, our method successfully pays more attention to entity mentions: ``\textit{she}'', ``\textit{her}'', ``\textit{he}'', and ``\textit{his}''. All of them are key entity mentions for the predicted relation.
We also can observe that the mention-aware segment attention of our SMS can focus on the core segment ``\textit{her dad}'', which is highly related to given entity pairs and matches the predicted label ``\textit{per:children}''.
From the second example, we can see that the model learns additional information which is similar to target relation. 
The model not only successfully pays attention on entity mention ``\textit{SUBJ-PER}'' and core segment ``\textit{COO of}'', but also captures similar entity mention ``\textit{Sally Strebel}'' and segment ``\textit{CEO of}'' simultaneously.
The case study proves that the mention attention and mention-aware segment attention do capture crucial entity mention level and segment level features.


\section{Related Works}
\subsection{RE with Neural Networks}
In recent years, neural networks have been large-scale used in relation extraction (RE). 
\citet{zeng2014relation,nguyen-grishman-2015-relation,wang2016relation} employ convolutional neural networks (CNN) to extract lexical and sentence level features for RE.
\citet{zhang2015relation} employs bidirectional recurrent neural networks (RNN) to learn long-term features to tackle long-term relation problems in RE. And many models with different attention mechanisms were proposed \cite{zhou2016attention,zhang2017position,xiao2016semantic,wang2016relation,yu2019beyond}.
\citet{vu-etal-2016-combining,nayak-ng-2019-effective} combine CNN and RNN to extract multi-types features from input sentences.
Recently, \citet{verga-etal-2018-simultaneously,DBLP:conf/ijcai/LiuCWZLX20} employ new neural structure transformer to extract features for RE, which is based on self-attention and is robust and powerful. 

Different from previous sequence-based models, dependency-based models employ dependency parsing of input sentences to capture non-local syntactic relations. The use of dependency trees has been a trend in relation extraction \cite{xu2015classifying,cai-etal-2016-bidirectional,miwa2016end,song-etal-2018-n}.
\citet{peng2017cross} split the dependency graph into two directed graphs, then extended the tree LSTM model \cite{tai2015improved} based on these two graphs to learn the structure of syntax dependency.
\citet{zhang2018graph} first introduced graph neural network \cite{kipf2016semi} (GNN) into RE model for encoding featrues from dependency tree and proposed a pruning strategy to remove unnecessary components of dependency tree. 
\citet{guo2019attention} also proposed a model with a soft-pruning approach that can automatically learn how to selectively attend to the relevant sub-structures useful for relation extraction.

\subsection{RE with Pretrained Models}
With the development of pre-trained language models \cite{devlin2018BERT}, the performance of relation extraction has been highly improved. After that, many researches based on BERT were carried out.
Most of these works employ pre-trained language models in three ways for relation extraction: 1) design task-related tasks in pre-training stage to improve prior pattern \cite{zhang-etal-2019-ernie,joshi-etal-2020-spanBERT,baldini-soares-etal-2019-matching,li2020downstream,peng-etal-2020-learning,yamada-etal-2020-luke}; 2) introduce external knowledge (e.g. knowledge graph and wiki data) into fine-tuning or pre-training stages \cite{peters-etal-2019-knowledge,baldini-soares-etal-2019-matching,wang2020kepler,wang2020kadapter,yamada-etal-2020-luke}; 3) employ representation from pre-trained language models and stack some neural structure over it \cite{8995204,alt_improving_2019,wang-etal-2019-extracting,wu2019enriching,shi2019simple,pmlr-v101-zhao19a,xue2020gdpnet,chen2020efficient}.
There are also some special methods with pre-trained language models \cite{li-etal-2019-entity,DBLP:conf/ijcai/ZhaoYCL20}. They convert relation classification tasks into machine reading comprehension tasks.
However, most of them is time-consuming or resource-consuming due to the require of external knowledge and the pre-train stage.


\section{Conclusion and Limitations}
In this paper, we analyze previous typical works and empirically focus on three granularity features: entity mention level, segment level and sentence level. Based on the hierarchical structure between entity mention level and segment level feature, we propose a multi-granularity hierarchical feature extractor for relation extraction, which does not need any external knowledge or tools. We evaluate our method with different encoders and results on three public benchmarks show that our method can bring outstanding improvement for them. 

The structure of our model make it not easy to apply on multi-relation extraction tasks. In the future, we will focus on how to extend our method to longer input tasks and multi-relation extraction tasks (e.g. Document Level Relation Extraction). Besides, we will also investigate what makes graph structure effective in relation extraction tasks and why our method can obtain better results than them.

\section*{Acknowledgements}
This work was supported in part by the National Natural Science Foundation of China (Grant Nos.U1636211, 61672081,61370126), the 2020 Tencent Wechat Rhino-Bird Focused Research Program, and the Fund of the State Key Laboratory of Software Development Environment (Grant No. SKLSDE-2021ZX-18).


\bibliography{anthology,custom}

\begin{thebibliography}{52}
\expandafter\ifx\csname natexlab\endcsname\relax\def\natexlab#1{#1}\fi

\bibitem[{Kim()}]{Kim2010}


\bibitem[{Alt et~al.(2020)Alt, Gabryszak, and Hennig}]{DBLP:conf/acl/AltGH20a}
Christoph Alt, Aleksandra Gabryszak, and Leonhard Hennig. 2020.
\newblock \href {https://doi.org/10.18653/v1/2020.acl-main.142} {{TACRED}
  revisited: {A} thorough evaluation of the {TACRED} relation extraction task}.
\newblock In \emph{Proceedings of the 58th Annual Meeting of the Association
  for Computational Linguistics, {ACL} 2020, Online, July 5-10, 2020}, pages
  1558--1569. Association for Computational Linguistics.

\bibitem[{Alt et~al.(2019)Alt, H\"{u}bner, and Hennig}]{alt_improving_2019}
Christoph Alt, Marc H\"{u}bner, and Leonhard Hennig. 2019.
\newblock \href {https://openreview.net/forum?id=BJgrxbqp67} {Improving
  relation extraction by pre-trained language representations}.
\newblock In \emph{Proceedings of AKBC 2019}.

\bibitem[{Baldini~Soares et~al.(2019)Baldini~Soares, FitzGerald, Ling, and
  Kwiatkowski}]{baldini-soares-etal-2019-matching}
Livio Baldini~Soares, Nicholas FitzGerald, Jeffrey Ling, and Tom Kwiatkowski.
  2019.
\newblock \href {https://doi.org/10.18653/v1/P19-1279} {Matching the blanks:
  Distributional similarity for relation learning}.
\newblock In \emph{Proceedings of the 57th Annual Meeting of the Association
  for Computational Linguistics}, pages 2895--2905, Florence, Italy.
  Association for Computational Linguistics.

\bibitem[{Cai et~al.(2016)Cai, Zhang, and Wang}]{cai-etal-2016-bidirectional}
Rui Cai, Xiaodong Zhang, and Houfeng Wang. 2016.
\newblock \href {https://doi.org/10.18653/v1/P16-1072} {Bidirectional recurrent
  convolutional neural network for relation classification}.
\newblock In \emph{Proceedings of the 54th Annual Meeting of the Association
  for Computational Linguistics (Volume 1: Long Papers)}, pages 756--765,
  Berlin, Germany. Association for Computational Linguistics.

\bibitem[{Chen et~al.(2020)Chen, Hoehndorf, Elhoseiny, and
  Zhang}]{chen2020efficient}
Jun Chen, Robert Hoehndorf, Mohamed Elhoseiny, and Xiangliang Zhang. 2020.
\newblock \href {http://arxiv.org/abs/2004.03636} {Efficient long-distance
  relation extraction with dg-spanbert}.

\bibitem[{Chowdhury and Lavelli(2012)}]{chowdhury-lavelli-2012-combining}
Md. Faisal~Mahbub Chowdhury and Alberto Lavelli. 2012.
\newblock \href {https://aclanthology.org/E12-1043} {Combining tree structures,
  flat features and patterns for biomedical relation extraction}.
\newblock In \emph{Proceedings of the 13th Conference of the {E}uropean Chapter
  of the Association for Computational Linguistics}, pages 420--429, Avignon,
  France. Association for Computational Linguistics.

\bibitem[{Devlin et~al.(2019)Devlin, Chang, Lee, and
  Toutanova}]{devlin2018BERT}
Jacob Devlin, Ming{-}Wei Chang, Kenton Lee, and Kristina Toutanova. 2019.
\newblock \href {https://doi.org/10.18653/v1/n19-1423} {{BERT:} pre-training of
  deep bidirectional transformers for language understanding}.
\newblock In \emph{Proceedings of the 2019 Conference of the North American
  Chapter of the Association for Computational Linguistics: Human Language
  Technologies, {NAACL-HLT} 2019, Minneapolis, MN, USA, June 2-7, 2019, Volume
  1 (Long and Short Papers)}, pages 4171--4186. Association for Computational
  Linguistics.

\bibitem[{Fader et~al.(2011)Fader, Soderland, and
  Etzioni}]{fader-etal-2011-identifying}
Anthony Fader, Stephen Soderland, and Oren Etzioni. 2011.
\newblock \href {https://www.aclweb.org/anthology/D11-1142} {Identifying
  relations for open information extraction}.
\newblock In \emph{Proceedings of the 2011 Conference on Empirical Methods in
  Natural Language Processing}, pages 1535--1545, Edinburgh, Scotland, UK.
  Association for Computational Linguistics.

\bibitem[{Guo et~al.(2019)Guo, Zhang, and Lu}]{guo2019attention}
Zhijiang Guo, Yan Zhang, and Wei Lu. 2019.
\newblock \href {https://doi.org/10.18653/v1/P19-1024} {Attention guided graph
  convolutional networks for relation extraction}.
\newblock In \emph{Proceedings of the 57th Annual Meeting of the Association
  for Computational Linguistics}, pages 241--251, Florence, Italy. Association
  for Computational Linguistics.

\bibitem[{Hendrickx et~al.(2010)Hendrickx, Kim, Kozareva, Nakov,
  {\'O}~S{\'e}aghdha, Pad{\'o}, Pennacchiotti, Romano, and
  Szpakowicz}]{hendrickx-etal-2010-semeval}
Iris Hendrickx, Su~Nam Kim, Zornitsa Kozareva, Preslav Nakov, Diarmuid
  {\'O}~S{\'e}aghdha, Sebastian Pad{\'o}, Marco Pennacchiotti, Lorenza Romano,
  and Stan Szpakowicz. 2010.
\newblock \href {https://www.aclweb.org/anthology/S10-1006} {{S}em{E}val-2010
  task 8: Multi-way classification of semantic relations between pairs of
  nominals}.
\newblock In \emph{Proceedings of the 5th International Workshop on Semantic
  Evaluation}, pages 33--38, Uppsala, Sweden. Association for Computational
  Linguistics.

\bibitem[{Ji and Grishman(2011)}]{ji-grishman-2011-knowledge}
Heng Ji and Ralph Grishman. 2011.
\newblock \href {https://www.aclweb.org/anthology/P11-1115} {Knowledge base
  population: Successful approaches and challenges}.
\newblock In \emph{Proceedings of the 49th Annual Meeting of the Association
  for Computational Linguistics: Human Language Technologies}, pages
  1148--1158, Portland, Oregon, USA. Association for Computational Linguistics.

\bibitem[{Joshi et~al.(2020)Joshi, Chen, Liu, Weld, Zettlemoyer, and
  Levy}]{joshi-etal-2020-spanBERT}
Mandar Joshi, Danqi Chen, Yinhan Liu, Daniel~S. Weld, Luke Zettlemoyer, and
  Omer Levy. 2020.
\newblock \href {https://doi.org/10.1162/tacl_a_00300} {{S}pan{BERT}: Improving
  pre-training by representing and predicting spans}.
\newblock \emph{Transactions of the Association for Computational Linguistics},
  8:64--77.

\bibitem[{Kipf and Welling(2017)}]{kipf2016semi}
Thomas~N. Kipf and Max Welling. 2017.
\newblock \href {https://openreview.net/forum?id=SJU4ayYgl} {Semi-supervised
  classification with graph convolutional networks}.
\newblock In \emph{5th International Conference on Learning Representations,
  {ICLR} 2017, Toulon, France, April 24-26, 2017, Conference Track
  Proceedings}. OpenReview.net.

\bibitem[{Li and Tian(2020)}]{li2020downstream}
Cheng Li and Ye~Tian. 2020.
\newblock \href {http://arxiv.org/abs/2004.03786} {Downstream model design of
  pre-trained language model for relation extraction task}.

\bibitem[{Li et~al.(2019)Li, Yin, Sun, Li, Yuan, Chai, Zhou, and
  Li}]{li-etal-2019-entity}
Xiaoya Li, Fan Yin, Zijun Sun, Xiayu Li, Arianna Yuan, Duo Chai, Mingxin Zhou,
  and Jiwei Li. 2019.
\newblock \href {https://doi.org/10.18653/v1/P19-1129} {Entity-relation
  extraction as multi-turn question answering}.
\newblock In \emph{Proceedings of the 57th Annual Meeting of the Association
  for Computational Linguistics}, pages 1340--1350, Florence, Italy.
  Association for Computational Linguistics.

\bibitem[{Liu et~al.(2020)Liu, Chen, Wang, Zhang, Li, and
  Xu}]{DBLP:conf/ijcai/LiuCWZLX20}
Jie Liu, Shaowei Chen, Bingquan Wang, Jiaxin Zhang, Na~Li, and Tong Xu. 2020.
\newblock \href {https://doi.org/10.24963/ijcai.2020/524} {Attention as
  relation: Learning supervised multi-head self-attention for relation
  extraction}.
\newblock In \emph{Proceedings of the Twenty-Ninth International Joint
  Conference on Artificial Intelligence, {IJCAI} 2020}, pages 3787--3793.
  ijcai.org.

\bibitem[{Miwa and Bansal(2016)}]{miwa2016end}
Makoto Miwa and Mohit Bansal. 2016.
\newblock \href {https://doi.org/10.18653/v1/P16-1105} {End-to-end relation
  extraction using {LSTM}s on sequences and tree structures}.
\newblock In \emph{Proceedings of the 54th Annual Meeting of the Association
  for Computational Linguistics (Volume 1: Long Papers)}, pages 1105--1116,
  Berlin, Germany. Association for Computational Linguistics.

\bibitem[{Nayak and Ng(2019)}]{nayak-ng-2019-effective}
Tapas Nayak and Hwee~Tou Ng. 2019.
\newblock \href {https://doi.org/10.18653/v1/K19-1056} {Effective attention
  modeling for neural relation extraction}.
\newblock In \emph{Proceedings of the 23rd Conference on Computational Natural
  Language Learning (CoNLL)}, pages 603--612, Hong Kong, China. Association for
  Computational Linguistics.

\bibitem[{Nguyen and Grishman(2015)}]{nguyen-grishman-2015-relation}
Thien~Huu Nguyen and Ralph Grishman. 2015.
\newblock \href {https://doi.org/10.3115/v1/W15-1506} {Relation extraction:
  Perspective from convolutional neural networks}.
\newblock In \emph{Proceedings of the 1st Workshop on Vector Space Modeling for
  Natural Language Processing}, pages 39--48, Denver, Colorado. Association for
  Computational Linguistics.

\bibitem[{Peng et~al.(2020)Peng, Gao, Han, Lin, Li, Liu, Sun, and
  Zhou}]{peng-etal-2020-learning}
Hao Peng, Tianyu Gao, Xu~Han, Yankai Lin, Peng Li, Zhiyuan Liu, Maosong Sun,
  and Jie Zhou. 2020.
\newblock \href {https://doi.org/10.18653/v1/2020.emnlp-main.298} {{L}earning
  from {C}ontext or {N}ames? {A}n {E}mpirical {S}tudy on {N}eural {R}elation
  {E}xtraction}.
\newblock In \emph{Proceedings of the 2020 Conference on Empirical Methods in
  Natural Language Processing (EMNLP)}, pages 3661--3672, Online. Association
  for Computational Linguistics.

\bibitem[{Peng et~al.(2017)Peng, Poon, Quirk, Toutanova, and
  Yih}]{peng2017cross}
Nanyun Peng, Hoifung Poon, Chris Quirk, Kristina Toutanova, and Wen{-}tau Yih.
  2017.
\newblock \href {https://transacl.org/ojs/index.php/tacl/article/view/1028}
  {Cross-sentence n-ary relation extraction with graph lstms}.
\newblock \emph{Trans. Assoc. Comput. Linguistics}, 5:101--115.

\bibitem[{Pennington et~al.(2014)Pennington, Socher, and
  Manning}]{pennington2014glove}
Jeffrey Pennington, Richard Socher, and Christopher Manning. 2014.
\newblock \href {https://doi.org/10.3115/v1/D14-1162} {{G}lo{V}e: Global
  vectors for word representation}.
\newblock In \emph{Proceedings of the 2014 Conference on Empirical Methods in
  Natural Language Processing ({EMNLP})}, pages 1532--1543, Doha, Qatar.
  Association for Computational Linguistics.

\bibitem[{Peters et~al.(2019{\natexlab{a}})Peters, Neumann, Logan, Schwartz,
  Joshi, Singh, and Smith}]{peters-etal-2019-knowledge}
Matthew~E. Peters, Mark Neumann, Robert Logan, Roy Schwartz, Vidur Joshi,
  Sameer Singh, and Noah~A. Smith. 2019{\natexlab{a}}.
\newblock \href {https://doi.org/10.18653/v1/D19-1005} {Knowledge enhanced
  contextual word representations}.
\newblock In \emph{Proceedings of the 2019 Conference on Empirical Methods in
  Natural Language Processing and the 9th International Joint Conference on
  Natural Language Processing (EMNLP-IJCNLP)}, pages 43--54, Hong Kong, China.
  Association for Computational Linguistics.

\bibitem[{Peters et~al.(2019{\natexlab{b}})Peters, Neumann, Logan, Schwartz,
  Joshi, Singh, and Smith}]{Peters2019KnowledgeEC}
Matthew~E. Peters, Mark Neumann, Robert Logan, Roy Schwartz, Vidur Joshi,
  Sameer Singh, and Noah~A. Smith. 2019{\natexlab{b}}.
\newblock \href {https://doi.org/10.18653/v1/D19-1005} {Knowledge enhanced
  contextual word representations}.
\newblock In \emph{Proceedings of the 2019 Conference on Empirical Methods in
  Natural Language Processing and the 9th International Joint Conference on
  Natural Language Processing (EMNLP-IJCNLP)}, pages 43--54, Hong Kong, China.
  Association for Computational Linguistics.

\bibitem[{Radford et~al.(2019)Radford, Wu, Child, Luan, Amodei, and
  Sutskever}]{radford2019language}
Alec Radford, Jeff Wu, Rewon Child, David Luan, Dario Amodei, and Ilya
  Sutskever. 2019.
\newblock Language models are unsupervised multitask learners.
\newblock \emph{OpenAI Blog}.

\bibitem[{Shi and Lin(2019)}]{shi2019simple}
Peng Shi and Jimmy Lin. 2019.
\newblock \href {http://arxiv.org/abs/1904.05255} {Simple bert models for
  relation extraction and semantic role labeling}.

\bibitem[{Song et~al.(2018)Song, Zhang, Wang, and Gildea}]{song-etal-2018-n}
Linfeng Song, Yue Zhang, Zhiguo Wang, and Daniel Gildea. 2018.
\newblock \href {https://doi.org/10.18653/v1/D18-1246} {N-ary relation
  extraction using graph-state {LSTM}}.
\newblock In \emph{Proceedings of the 2018 Conference on Empirical Methods in
  Natural Language Processing}, pages 2226--2235, Brussels, Belgium.
  Association for Computational Linguistics.

\bibitem[{Tai et~al.(2015)Tai, Socher, and Manning}]{tai2015improved}
Kai~Sheng Tai, Richard Socher, and Christopher~D. Manning. 2015.
\newblock \href {https://doi.org/10.3115/v1/P15-1150} {Improved semantic
  representations from tree-structured long short-term memory networks}.
\newblock In \emph{Proceedings of the 53rd Annual Meeting of the Association
  for Computational Linguistics and the 7th International Joint Conference on
  Natural Language Processing (Volume 1: Long Papers)}, pages 1556--1566,
  Beijing, China. Association for Computational Linguistics.

\bibitem[{{Tao} et~al.(2019){Tao}, {Luo}, {Wang}, and {Xu}}]{8995204}
Q.~{Tao}, X.~{Luo}, H.~{Wang}, and R.~{Xu}. 2019.
\newblock \href {https://doi.org/10.1109/ICTAI.2019.00227} {Enhancing relation
  extraction using syntactic indicators and sentential contexts}.
\newblock In \emph{2019 IEEE 31st International Conference on Tools with
  Artificial Intelligence (ICTAI)}, pages 1574--1580.

\bibitem[{Verga et~al.(2018)Verga, Strubell, and
  McCallum}]{verga-etal-2018-simultaneously}
Patrick Verga, Emma Strubell, and Andrew McCallum. 2018.
\newblock \href {https://doi.org/10.18653/v1/N18-1080} {Simultaneously
  self-attending to all mentions for full-abstract biological relation
  extraction}.
\newblock In \emph{Proceedings of the 2018 Conference of the North {A}merican
  Chapter of the Association for Computational Linguistics: Human Language
  Technologies, Volume 1 (Long Papers)}, pages 872--884, New Orleans,
  Louisiana. Association for Computational Linguistics.

\bibitem[{Vu et~al.(2016)Vu, Adel, Gupta, and
  Sch{\"u}tze}]{vu-etal-2016-combining}
Ngoc~Thang Vu, Heike Adel, Pankaj Gupta, and Hinrich Sch{\"u}tze. 2016.
\newblock \href {https://doi.org/10.18653/v1/N16-1065} {Combining recurrent and
  convolutional neural networks for relation classification}.
\newblock In \emph{Proceedings of the 2016 Conference of the North {A}merican
  Chapter of the Association for Computational Linguistics: Human Language
  Technologies}, pages 534--539, San Diego, California. Association for
  Computational Linguistics.

\bibitem[{Wang et~al.(2019)Wang, Tan, Yu, Chang, Wang, Xu, Guo, and
  Potdar}]{wang-etal-2019-extracting}
Haoyu Wang, Ming Tan, Mo~Yu, Shiyu Chang, Dakuo Wang, Kun Xu, Xiaoxiao Guo, and
  Saloni Potdar. 2019.
\newblock \href {https://doi.org/10.18653/v1/P19-1132} {Extracting
  multiple-relations in one-pass with pre-trained transformers}.
\newblock In \emph{Proceedings of the 57th Annual Meeting of the Association
  for Computational Linguistics}, pages 1371--1377, Florence, Italy.
  Association for Computational Linguistics.

\bibitem[{Wang et~al.(2016)Wang, Cao, de~Melo, and Liu}]{wang2016relation}
Linlin Wang, Zhu Cao, Gerard de~Melo, and Zhiyuan Liu. 2016.
\newblock \href {https://doi.org/10.18653/v1/P16-1123} {Relation classification
  via multi-level attention {CNN}s}.
\newblock In \emph{Proceedings of the 54th Annual Meeting of the Association
  for Computational Linguistics (Volume 1: Long Papers)}, pages 1298--1307,
  Berlin, Germany. Association for Computational Linguistics.

\bibitem[{Wang et~al.(2020{\natexlab{a}})Wang, Tang, Duan, Wei, Huang, ji, Cao,
  Jiang, and Zhou}]{wang2020kadapter}
Ruize Wang, Duyu Tang, Nan Duan, Zhongyu Wei, Xuanjing Huang, Jianshu ji,
  Guihong Cao, Daxin Jiang, and Ming Zhou. 2020{\natexlab{a}}.
\newblock \href {http://arxiv.org/abs/2002.01808} {K-adapter: Infusing
  knowledge into pre-trained models with adapters}.

\bibitem[{Wang et~al.(2020{\natexlab{b}})Wang, Gao, Zhu, Zhang, Liu, Li, and
  Tang}]{wang2020kepler}
Xiaozhi Wang, Tianyu Gao, Zhaocheng Zhu, Zhengyan Zhang, Zhiyuan Liu, Juanzi
  Li, and Jian Tang. 2020{\natexlab{b}}.
\newblock \href {http://arxiv.org/abs/1911.06136} {Kepler: A unified model for
  knowledge embedding and pre-trained language representation}.

\bibitem[{Wu and He(2019{\natexlab{a}})}]{wu-etal-2019-r-bert}
Shanchan Wu and Yifan He. 2019{\natexlab{a}}.
\newblock \href {https://doi.org/10.1145/3357384.3358119} {Enriching
  pre-trained language model with entity information for relation
  classification}.
\newblock In \emph{Proceedings of the 28th ACM International Conference on
  Information and Knowledge Management}, CIKM '19, page 2361–2364, New York,
  NY, USA. Association for Computing Machinery.

\bibitem[{Wu and He(2019{\natexlab{b}})}]{wu2019enriching}
Shanchan Wu and Yifan He. 2019{\natexlab{b}}.
\newblock \href {http://arxiv.org/abs/1905.08284} {Enriching pre-trained
  language model with entity information for relation classification}.

\bibitem[{Xiao and Liu(2016)}]{xiao2016semantic}
Minguang Xiao and Cong Liu. 2016.
\newblock \href {https://www.aclweb.org/anthology/C16-1119} {Semantic relation
  classification via hierarchical recurrent neural network with attention}.
\newblock In \emph{Proceedings of {COLING} 2016, the 26th International
  Conference on Computational Linguistics: Technical Papers}, pages 1254--1263,
  Osaka, Japan. The COLING 2016 Organizing Committee.

\bibitem[{Xu et~al.(2015)Xu, Mou, Li, Chen, Peng, and Jin}]{xu2015classifying}
Yan Xu, Lili Mou, Ge~Li, Yunchuan Chen, Hao Peng, and Zhi Jin. 2015.
\newblock \href {https://doi.org/10.18653/v1/D15-1206} {Classifying relations
  via long short term memory networks along shortest dependency paths}.
\newblock In \emph{Proceedings of the 2015 Conference on Empirical Methods in
  Natural Language Processing}, pages 1785--1794, Lisbon, Portugal. Association
  for Computational Linguistics.

\bibitem[{Xue et~al.(2020)Xue, Sun, Zhang, and Chng}]{xue2020gdpnet}
Fuzhao Xue, Aixin Sun, Hao Zhang, and Eng~Siong Chng. 2020.
\newblock \href {http://arxiv.org/abs/2012.06780} {Gdpnet: Refining latent
  multi-view graph for relation extraction}.

\bibitem[{Yamada et~al.(2020)Yamada, Asai, Shindo, Takeda, and
  Matsumoto}]{yamada-etal-2020-luke}
Ikuya Yamada, Akari Asai, Hiroyuki Shindo, Hideaki Takeda, and Yuji Matsumoto.
  2020.
\newblock \href {https://doi.org/10.18653/v1/2020.emnlp-main.523} {{LUKE}: Deep
  contextualized entity representations with entity-aware self-attention}.
\newblock In \emph{Proceedings of the 2020 Conference on Empirical Methods in
  Natural Language Processing (EMNLP)}, pages 6442--6454, Online. Association
  for Computational Linguistics.

\bibitem[{Yu et~al.(2019)Yu, Zhang, Liu, Wang, Li, and Li}]{yu2019beyond}
Bowen Yu, Zhenyu Zhang, Tingwen Liu, Bin Wang, Sujian Li, and Quangang Li.
  2019.
\newblock \href {https://doi.org/10.24963/ijcai.2019/750} {Beyond word
  attention: Using segment attention in neural relation extraction}.
\newblock In \emph{Proceedings of the Twenty-Eighth International Joint
  Conference on Artificial Intelligence, {IJCAI-19}}, pages 5401--5407.
  International Joint Conferences on Artificial Intelligence Organization.

\bibitem[{Yu et~al.(2017)Yu, Yin, Hasan, dos Santos, Xiang, and
  Zhou}]{yu-etal-2017-improved}
Mo~Yu, Wenpeng Yin, Kazi~Saidul Hasan, Cicero dos Santos, Bing Xiang, and Bowen
  Zhou. 2017.
\newblock \href {https://doi.org/10.18653/v1/P17-1053} {Improved neural
  relation detection for knowledge base question answering}.
\newblock In \emph{Proceedings of the 55th Annual Meeting of the Association
  for Computational Linguistics (Volume 1: Long Papers)}, pages 571--581,
  Vancouver, Canada. Association for Computational Linguistics.

\bibitem[{Zeng et~al.(2014)Zeng, Liu, Lai, Zhou, and Zhao}]{zeng2014relation}
Daojian Zeng, Kang Liu, Siwei Lai, Guangyou Zhou, and Jun Zhao. 2014.
\newblock \href {https://www.aclweb.org/anthology/C14-1220} {Relation
  classification via convolutional deep neural network}.
\newblock In \emph{Proceedings of {COLING} 2014, the 25th International
  Conference on Computational Linguistics: Technical Papers}, pages 2335--2344,
  Dublin, Ireland. Dublin City University and Association for Computational
  Linguistics.

\bibitem[{Zhang and Wang(2015)}]{zhang2015relation}
Dongxu Zhang and Dong Wang. 2015.
\newblock Relation classification via recurrent neural network.
\newblock \emph{arXiv preprint arXiv:1508.01006}.

\bibitem[{Zhang et~al.(2018)Zhang, Qi, and Manning}]{zhang2018graph}
Yuhao Zhang, Peng Qi, and Christopher~D. Manning. 2018.
\newblock \href {https://doi.org/10.18653/v1/D18-1244} {Graph convolution over
  pruned dependency trees improves relation extraction}.
\newblock In \emph{Proceedings of the 2018 Conference on Empirical Methods in
  Natural Language Processing}, pages 2205--2215, Brussels, Belgium.
  Association for Computational Linguistics.

\bibitem[{Zhang et~al.(2017)Zhang, Zhong, Chen, Angeli, and
  Manning}]{zhang2017position}
Yuhao Zhang, Victor Zhong, Danqi Chen, Gabor Angeli, and Christopher~D.
  Manning. 2017.
\newblock \href {https://doi.org/10.18653/v1/D17-1004} {Position-aware
  attention and supervised data improve slot filling}.
\newblock In \emph{Proceedings of the 2017 Conference on Empirical Methods in
  Natural Language Processing}, pages 35--45, Copenhagen, Denmark. Association
  for Computational Linguistics.

\bibitem[{Zhang et~al.(2019)Zhang, Han, Liu, Jiang, Sun, and
  Liu}]{zhang-etal-2019-ernie}
Zhengyan Zhang, Xu~Han, Zhiyuan Liu, Xin Jiang, Maosong Sun, and Qun Liu. 2019.
\newblock \href {https://doi.org/10.18653/v1/P19-1139} {{ERNIE}: Enhanced
  language representation with informative entities}.
\newblock In \emph{Proceedings of the 57th Annual Meeting of the Association
  for Computational Linguistics}, pages 1441--1451, Florence, Italy.
  Association for Computational Linguistics.

\bibitem[{Zhao et~al.(2020)Zhao, Yan, Cao, and Li}]{DBLP:conf/ijcai/ZhaoYCL20}
Tianyang Zhao, Zhao Yan, Yunbo Cao, and Zhoujun Li. 2020.
\newblock Asking effective and diverse questions: A machine reading
  comprehension based framework for joint entity-relation extraction.
\newblock In \emph{Proceedings of the Twenty-Ninth International Joint
  Conference on Artificial Intelligence, {IJCAI-20}}, pages 3948--3954.
  International Joint Conferences on Artificial Intelligence Organization.
\newblock Main track.

\bibitem[{Zhao et~al.(2019)Zhao, Wan, Gao, and Lin}]{pmlr-v101-zhao19a}
Yi~Zhao, Huaiyu Wan, Jianwei Gao, and Youfang Lin. 2019.
\newblock \href {http://proceedings.mlr.press/v101/zhao19a.html} {Improving
  relation classification by entity pair graph}.
\newblock In \emph{Proceedings of The Eleventh Asian Conference on Machine
  Learning}, volume 101 of \emph{Proceedings of Machine Learning Research},
  pages 1156--1171, Nagoya, Japan. PMLR.

\bibitem[{Zhou et~al.(2016)Zhou, Shi, Tian, Qi, Li, Hao, and
  Xu}]{zhou2016attention}
Peng Zhou, Wei Shi, Jun Tian, Zhenyu Qi, Bingchen Li, Hongwei Hao, and Bo~Xu.
  2016.
\newblock \href {https://doi.org/10.18653/v1/P16-2034} {Attention-based
  bidirectional long short-term memory networks for relation classification}.
\newblock In \emph{Proceedings of the 54th Annual Meeting of the Association
  for Computational Linguistics (Volume 2: Short Papers)}, pages 207--212,
  Berlin, Germany. Association for Computational Linguistics.

\end{thebibliography}
\bibliographystyle{acl_natbib}




\end{document}